\documentclass[runningheads]{llncs}
\usepackage{graphicx}
\usepackage{amsmath,amssymb} % define this before the line numbering.
\usepackage{color}

\usepackage{makecell}
\usepackage[noend]{algpseudocode}
\usepackage{wrapfig,booktabs}
\usepackage[numbers,sort&compress]{natbib}
\usepackage{color}
\usepackage[normalem]{ulem}
\newcommand{\cmark}{\ding{51}}%
\newcommand{\xmark}{\ding{55}}%
\usepackage[dvipsnames]{xcolor}
\usepackage[ruled,vlined]{algorithm2e}
\usepackage{float}
\usepackage{pifont}
\usepackage{graphicx}
\usepackage{optidef}
\usepackage{multirow}
\usepackage{mathtools}
\usepackage{dsfont}
\usepackage{comment}
\usepackage{enumitem,kantlipsum}
\usepackage{diagbox}
\usepackage{wrapfig,lipsum,booktabs}
\makeatletter
\@namedef{ver@everyshi.sty}{}
\makeatother
\newcommand\normx[1]{\left\Vert#1\right\Vert}

\newcommand{\ours}{\textsc{SOCKET}}
\def\eg{\emph{e.g}.}
\def\ie{\emph{i.e}.}

\begin{document}
\pagestyle{headings}
\mainmatter
\def\ECCV16SubNumber{5271}  % Insert your submission number here

\title{Cross-Modal Knowledge Transfer Without Task-Relevant Source Data} % Replace with your title

\titlerunning{SOCKET}

\author{Sk Miraj Ahmed\inst{1}\index{Ahmed, Sk Miraj} \and
Suhas Lohit\inst{2}\index{Lohit, Suhas} \and Kuan-Chuan Peng\inst{2}\index{Peng, Kuan-Chuan} \and Michael J. Jones\inst{2}\index{Jones, Michael J.} \and Amit K. Roy-Chowdhury\inst{1}\index{Roy-Chowdhury, Amit K.}
}
\authorrunning{Ahmed et al.}
\institute{University of California, Riverside, CA 92507, USA \\
\email{\{sahme047@,amitrc@ece.\}ucr.edu}
\and
Mitsubishi Electric Research Laboratories (MERL), Cambridge, MA 02139, USA\\
\url{https://www.merl.com} 
\email{\{slohit,kpeng,mjones\}@merl.com}
}

\maketitle
\begin{abstract}
Cost-effective depth and infrared sensors as alternatives to usual RGB sensors are now a reality, and have some advantages over RGB in domains like autonomous navigation and remote sensing. As such, building computer vision and deep learning systems for depth and infrared data are crucial. However, large labeled datasets for these modalities are still lacking. In such cases, transferring knowledge from a neural network trained on a well-labeled large dataset in the source modality (RGB) to a neural network that works on a target modality (depth, infrared, etc.) is of great value. For reasons like memory and privacy, it may not be possible to access the source data, and knowledge transfer needs to work with only the source models. We describe an effective solution, \ours: SOurce-free Cross-modal KnowledgE Transfer for this challenging task of transferring knowledge from one source modality to a different target modality without access to task-relevant source data. The framework reduces the modality gap using paired task-irrelevant data, as well as by matching the mean and variance of the target features with the batch-norm statistics that are present in the source models. We show through extensive experiments that our method significantly outperforms existing source-free methods for classification tasks which do not account for the modality gap.

\end{abstract}

\section{Introduction}\label{sec:intro}

Depth sensors like Kinect and RealSense, LIDAR for measuring point clouds directly, or high resolution infra-red sensors such as from FLIR, allow for expanding the range of applications of computer vision compared to using only visible wavelengths. Sensing depth directly can provide an approximate three-dimensional picture of the scene and thus improve the performance of applications like autonomous navigation, while sensing in the infra-red wavelengths can allow for easier pedestrian detection or better object detection in adverse atmospheric conditions like rain, fog, and smoke. These are just a few examples.

\begin{figure}[t]
\centering
\includegraphics[width=0.55\textwidth]{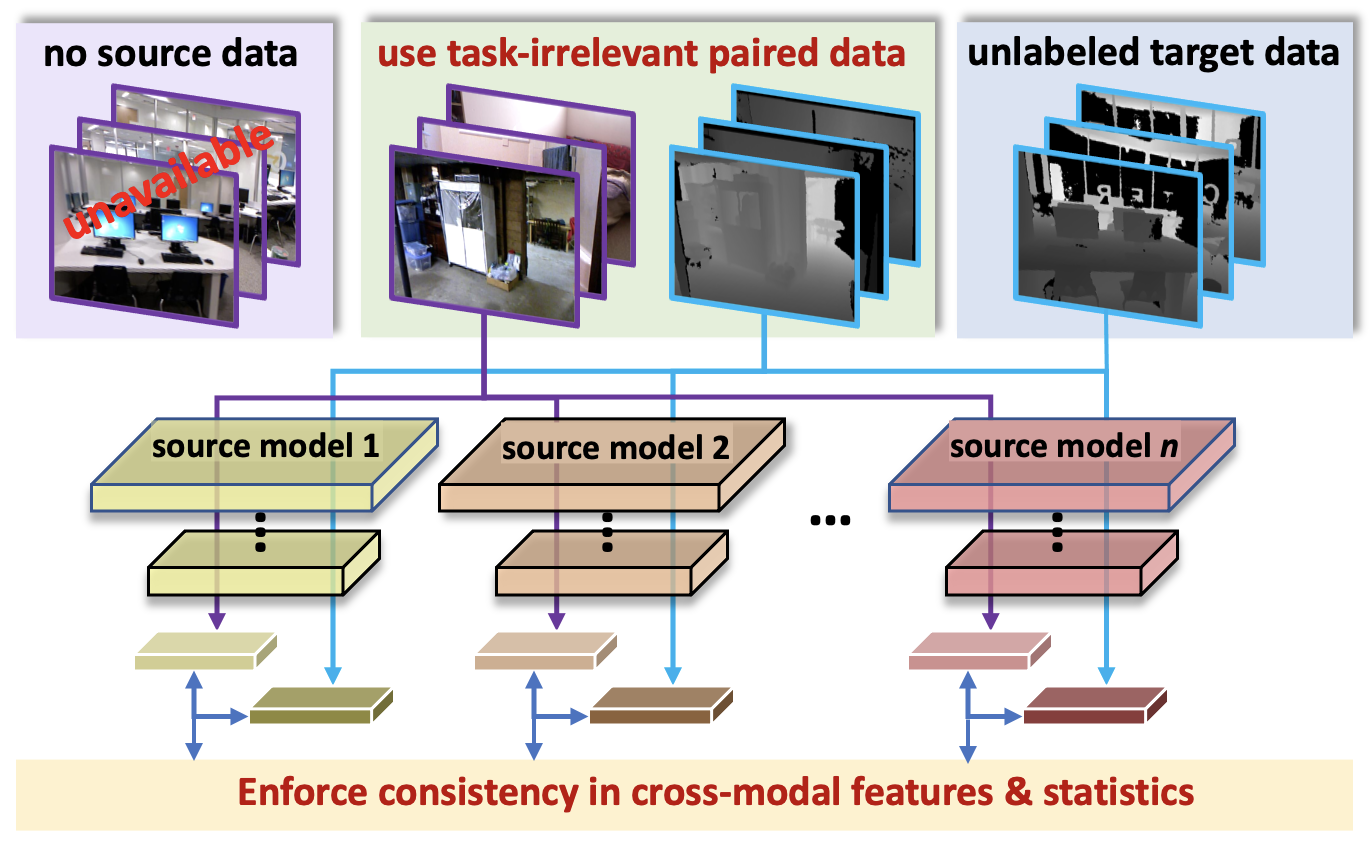}
\caption{\textbf{\ours:} We describe the problem of single/multi-source cross-modality knowledge transfer using no data used to train the source models. To effectively perform knowledge transfer, we minimize the modality gap by enforcing consistency of cross modal features on \textbf{task-irrelevant} paired data in feature space, and by \textbf{matching the distributions} of the unlabeled task-relevant features and the source features}
\label{fig:intro}
\end{figure}

Building computer vision applications using the now-straightforward supervised deep learning approach for modalities like depth and infrared needs large amounts of diverse labeled data. However, such large and diverse datasets do not exist for these modalities and the cost of building such datasets can be prohibitively high. In such cases, researchers have developed methods like knowledge distillation to transfer the knowledge from a model trained on a modality like RGB, where large amounts of labeled data are available, to the modality of interest like depth~\cite{gupta2016cross}.  

In contrast to prior work, we tackle a novel and challenging problem in the context of cross-modal knowledge transfer. We assume that we have access only to (a) the source models trained for the task of interest (TOI), and (b) unlabeled data in the target modality where we need to construct a model for the same TOI. The key aspect is that we assume we have \textbf{no access to any data in the source modality} for TOI. Such a problem setup is important in cases where memory and privacy considerations do not allow for sharing the training data from the source modality; only the trained models can be shared \cite{ahmed2020camera,liang2020we,ahmed2021unsupervised,perrot2015theoretical}. 

We develop \textbf{\ours}: \textbf{SO}urce-free \textbf{C}ross-modal \textbf{K}nowledg\textbf{E} \textbf{T}ransfer as an effective solution to this problem for bridging the gap between the source and target modalities. To this end, we show that employing an external dataset of source-target modality pairs, which are not relevant to TOI -- which we call Task-Irrelevant (TI) data -- can help in learning an effective target model by bringing the features of the two modalities closer. In addition to using TI data, we encourage matching the statistics of the features of the unlabeled target data -- which are Task-Relevant (TR) by definition -- with the statistics of the source data which are available to us from the normalization layers that are present in the trained source model.

We provide important empirical evidence showing that the modality-shift from a source modality like RGB to a target modality like depth can be much more challenging than a domain shift from one RGB dataset to another. This shows that the proposed framework is necessary to help minimize the modality gap, so as to make the knowledge transfer more effective. Based on the above ideas, we show that we can improve on existing state-of-the-art methods which were devised only for cross-domain setting in the same modality. We summarize our main contributions below:

\begin{enumerate}[label=\arabic*., leftmargin=*,topsep=0pt]
    \item We formulate a novel problem for knowledge transfer from a model trained for a source modality to a different target modality without any  access to task-relevant source data and when the target data is unlabeled.
    \item In order to bridge the gap between modalities, we propose a novel framework, \ours, for cross-modal knowledge transfer without access to source data (a) using an external task-irrelevant paired dataset, and (b) by matching the moments obtained from the normalization layers in the source models with the moments computed on the unlabeled target data.
    \item Extensive experiments on multiple datasets -- both for knowledge transfer from RGB to depth, and from RGB to IR, and both for single-source and multi-source cases -- show that SOCKET is useful in reducing the modality gap in the feature space and produces significantly better performance (improvement of as high as 12\% for some cases) over the existing source-free domain adaptation baselines which do not account for the modality difference between the source and target modalities.
    \item We also show empirically that, for the datasets of interest, the problem of knowledge transfer between modalities like RGB and depth is harder than domain shifts in the same modality such as sensor changes and viewpoint shifts, considered previously in literature.
\end{enumerate}

\section{Related work}\label{sec:related_work}
\textbf{Cross-modal distillation methods.} 
Cross-modal knowledge distillation (CMKD) methods aim to learn representations for a modality which does not have a large amount of labeled data from a large labeled dataset of another modality \cite{gupta2016cross}. These methods have been used for a variety of practical computer vision and learning tasks \cite{thoker2019cross,dai2021learning,garcia2019dmcl,wang2021cross}. Most of these works assume access to task-relevant paired data across modalities \cite{gupta2016cross,sayed2018cross,garcia2019dmcl,hoffman2016cross}. A recent line of work relaxed this assumption in the context of domain generalization, where one does not have access to the Task-Relevant paired data on the target domain but has access to them for the source domain \cite{zhao2020knowledge}. There also exist some works regarding domain translation across modalities for better classification of indoor scenes \cite{ferreri2021translate,du2019translate,ayub2019centroid}. However these methods consider UDA across domains, where the target domain has unlabeled RGB-D pairs instead of a single modality. All of the above works either utilize the Task-Relevant paired data for cross modal knowledge transfer \cite{gupta2016cross}, or consider cross modal paired data as a domain \cite{zhao2020knowledge,ferreri2021translate}. There are also works in zero-shot domain adaptation that utilize external task-irrelevant paired data \cite{peng2018zero} but need access to the source data. Our work takes steps to allow for different source and target modalities, and can perform effective knowledge transfer without access to the TR paired data between source and target. 

\begin{table*}[ht]

\centering
\caption{We compare the proposed work \ours~with existing problem settings in literature for knowledge transfer across different domains and modalities. The competitive settings described in this table are: (1) UDA (Unsupervised Domain Adaptation), DT (Domain Translation) \cite{hsu2020progressive,tzeng2017adversarial,paul2020domain,hoffman2018cycada,ferreri2021translate,du2019translate,ayub2019centroid} [$\mathcal{C}_1$], (2) MSDA (Multi-source domain adaptation) \cite{peng2019moment} [$\mathcal{C}_2$], (3) SFDA (Source free single source DA) \cite{liang2020we,yang2021generalized,yang2021exploiting,yang2020casting,agarwal2022unsupervised,liang2021source} [$\mathcal{C}_3$], (4) MSFDA (Source free multi-source DA) \cite{ahmed2021unsupervised} [$\mathcal{C}_4$], (5) CMKD (Cross modal knowledge distillation) \cite{gupta2016cross,thoker2019cross,dai2021learning,garcia2019dmcl} [$\mathcal{C}_5$], and (6) ZDDA (Zero shot DA) \cite{peng2018zero} [$\mathcal{C}_6$], respectively. We group citations into [$\mathcal{C}_1]$ to  $[\mathcal{C}_6$] based on problem settings. Only \ours~allows cross-modal knowledge transfer from multiple sources without any access to relevant source training data for an unlabeled target dataset of a different modality
}
\resizebox{\textwidth}{!}{
\begin{tabular}{l c c c c c c c}
\toprule
\backslashbox{Property}{Problem setting} &UDA+DT [$\mathcal{C}_1$] &MSDA [$\mathcal{C}_2$] &SFDA [$\mathcal{C}_3$] &MSFDA [$\mathcal{C}_4$] &CMKD [$\mathcal{C}_5$] &ZDDA [$\mathcal{C}_6$] &\ours \\ \midrule
Multiple sources
& \color{red}\xmark
& \color{ForestGreen}\cmark    
& \color{red}\xmark 
& \color{red}\xmark 
& \color{red}\xmark 
& \color{red}\xmark
& \color{ForestGreen}\cmark \\
No source data
& \color{red}\xmark
& \color{red}\xmark
& \color{ForestGreen}\cmark 
& \color{ForestGreen}\cmark 
& \color{red}\xmark 
& \color{red}\xmark
&  \color{ForestGreen}\cmark\\
Unlabeled target data
& \color{ForestGreen}\cmark
& \color{ForestGreen}\cmark
& \color{red}\xmark
& \color{ForestGreen}\cmark 
& \color{red}\xmark 
& \color{red}\xmark
& \color{ForestGreen}\cmark\\
Different target modality
& \color{red}\xmark
& \color{red}\xmark
& \color{red}\xmark
& \color{red}\xmark 
& \color{ForestGreen}\cmark 
& \color{ForestGreen}\cmark 
& \color{ForestGreen}\cmark\\
Usage of Task-Irrelevant Data
& \color{red}\xmark
& \color{red}\xmark
& \color{red}\xmark
& \color{red}\xmark 
& \color{red}\xmark 
& \color{ForestGreen}\cmark 
& \color{ForestGreen}\cmark\\
\bottomrule \\
\end{tabular}
}

\label{tab:related}
\end{table*}
\noindent
\textbf{Unsupervised domain adaptation methods without source data.} 
Most UDA methods that have been used for a wide variety of tasks \cite{hsu2020progressive,tzeng2017adversarial,paul2020domain,hoffman2018cycada} need access to the source data while adapting to a new target domain \cite{ganin2016domain,peng2019moment}. To combat the storage or privacy issue regarding the source data, a new line of work named Hypothesis Transfer Learning (HTL) \cite{ahmed2020camera,perrot2015theoretical} has emerged recently, where one has access only to the trained source model instead of the source data \cite{ahmed2021unsupervised,liang2020we}. Here, people have explored adapting target domain data, which has limited labels \cite{ahmed2020camera} or no labels at all \cite{liang2020we} in the presence of both single source \cite{liang2020we,yang2021generalized,yang2021exploiting} or multiple source models \cite{ahmed2021unsupervised}. \cite{liang2020we,liang2021source} adapts a single source model to an unlabeled target domain via information maximization and an iterative self-supervised pseudo-label based cross entropy loss. \cite{yang2021generalized} ensured that the adapted source model performs well, both on source and target domains, while \cite{yang2021exploiting} proposed a source free domain adaptation (SFDA) method by encouraging label consistency among local target features. \cite{yang2020casting} proposed to add an extra classifier for refinement of the source decision boundary, while \cite{agarwal2022unsupervised} proposed a more robust adaptation method which works well in the presence of noisy pseudo-labels.
The authors in \cite{ahmed2021unsupervised} proposed fusion of multiple source models with appropriate weights so as to minimize the effect of negative transfer, which we refer to as multiple source free domain adaptation (MSFDA) in Table~\ref{tab:related}. Both these methods do not work well in a regime where the unlabeled target set is from a different modality than the source. We solve this problem by modality gap reduction via feature matching of the task-irrelevant external data, as well as data statistics matching between the source and target modalities.

Table \ref{tab:related} summarizes the related work and compares them with SOCKET.

\section{Problem setup and notation}

We address the problem of source-data free cross-modality knowledge transfer by devising specialized loss functions that help reduce the gap between source and target modality features. We focus on the task of classification where both the source and target data belong to the same $N$ classes. Let us consider that we have $n$ source models of the same modality (\eg, RGB). We denote the trained source classifiers as $\{\mathcal{F}_{S_k}^{m_S}\}_{k=1}^n$ , where $S_k$ denotes the $k$-th source model and $m_S$ represents the modality on which the source models were trained. The source models are denoted as $\mathcal{F}_{S_k}^{m_S}$ which are trained models that map images from the source modality distribution  $\mathcal{X}_{S_k}^{m_S}$ to probability distribution over the classes. $\{x_{S_k}^i,y_{S_k}^i\}_{i=1}^{n_k} \sim \mathcal{X}_{S_k}^{m_S}$ are the data on which the $k$-th source model was trained, $n_k$ being the number of training data points corresponding to the $k$-th source. In our problem setting, at the time of knowledge transfer to the target modality, the source data are unavailable for all the sources.

\begin{figure}[t]
\centering
\includegraphics[width=1\textwidth]{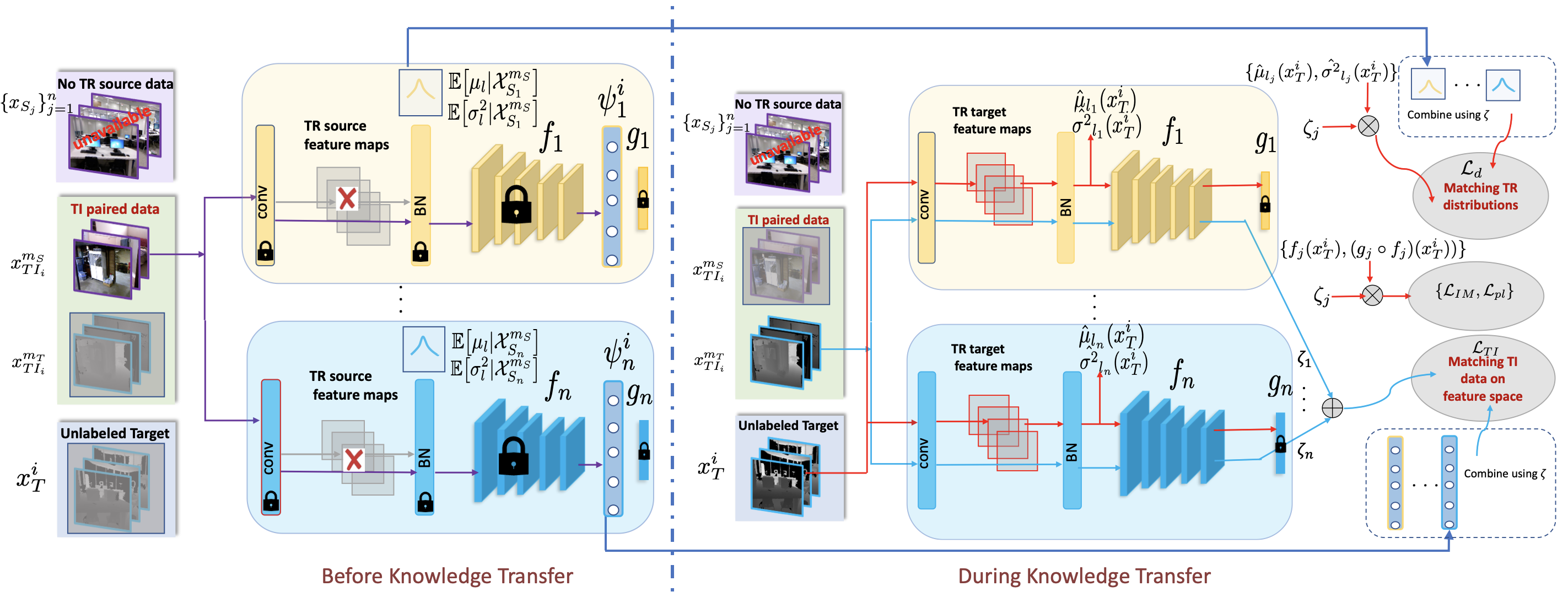}

\caption{\textbf{\ours~ description:} Our framework can be split into two parts: (i) Before Knowledge Transfer (left): We freeze the source models and pass the task-irrelevant (TI) source data through the source feature encoders to extract the TI source features. As task-relevant (TR) source feature maps are not available, we extract the stored moments of its distribution from the BN layers. (ii) During Knowledge Transfer (right): We freeze only the classification layers and feed the TI and unlabeled TR target data  through the models to get batch-wise TI target features and the TR target moments, respectively, which we match with pre-extracted source features and moments to jointly train all the feature encoders along with the mixing weights, $\zeta_k$'s. The final target model is the optimal linear combination of the updated source models}

\label{framework}
\end{figure}

We also have access to an unlabeled dataset in the target modality $\{x_{T}^i\}_{i=1}^{n_T} \sim \mathcal{X}_{T}^{m_T}$, where $n_T$ is the number of target samples. Note that the target modality, $m_T$, is different from the source modality. Traditional source free UDA methods try to mitigate domain shift by adapting the source models to unlabeled target data that belong to the same modality \cite{liang2020we, ahmed2021unsupervised}. As we will show, applying these methods directly to the cross-modal setting results in poor performance. Hence, we propose to solve this problem using two novel losses as regularization terms which minimize the modality gap between source and target modalities. Our goal is to learn a target classifier  $\mathcal{F}_{T}^{m_T}$, that adapts well on a target distribution obtained from a different sensor modality (\eg, depth or NIR). 

To train $\mathcal{F}_{T}^{m_T}$, we employ (a) methods that enable learning feature embeddings for the target modality that closely match with the source modality embeddings, which we group under modality-specific losses, since it bridges the gap between two different modalities; (b) modality-agnostic loss terms which operate only on the unlabeled target data and do not take into account shift in modality.

We split each of the source models into two blocks -- \textit{feature encoder} and \textit{classifier}. For the $k$-th source model, we denote these blocks as $f_{k}$ and $g_k$, respectively. The function $f_{k}: \mathbb{R}^{H \times W} \rightarrow \mathbb{R}^{\eta}$ maps the input image to an $\eta$ dimensional feature vector and $g_k: \mathbb{R}^{\eta} \rightarrow \mathbb{R}^N$ maps those features to the probability distributions over the $N$ classes, the maximum of which is treated as the classifier prediction. We can thus write $\mathcal{F}_{S_k}^{m_S} = {g}_k \circ f_{k}$ , where ``$\circ$" is function composition. Since the classifier layer $g_k$ contains the information about unseen $k$-th source domain distribution, following the protocol of \cite{ahmed2021unsupervised}, we freeze all the $g_k$'s and train the target specific feature encoders by optimizing over all $f_k$'s.

\section{Cross-Modal Feature Alignment}
Traditional source free UDA methods~\cite{liang2020we, ahmed2021unsupervised} use domain specific but modality-agnostic losses which do not help in reducing the feature distance between the source and target modalities. In order to train the target model, $\mathcal{F}_{T}^{m_T}$, with reduced modality-gap, we propose \ours, which uses \textit{task-irrelevant feature matching} and \textit{task-relevant distribution matching} which are described next. 

\subsection{Task-irrelevant feature matching}

Capturing the mapping between two modalities effectively requires lots of paired data from both modalities \cite{bridle1992unsupervised}. For our task of interest, we do not have task relevant (TR) data on the source side. As a result, it is not possible to match the target modality with the source modality by using the data from task relevant classes directly. Hence, we propose to use \textbf{Task-Irrelevant (TI) paired data} from both modalities to reduce modality gap. TI data contain only classes that are completely \textbf{disjoint} from the TR classes and can be from any external dataset. For modalities like RGB-depth and RGB-IR, we can access a large amount of paired TI data that contain classes with no privacy concerns, which are available in public datasets or can be collected using multi-modal sensors. Moreover there are many real world applications where pairwise TI data can be collected and used beyond RGB-D or RGB-IR, such as autonomous driving, adpatation of LiDAR data, medical applications \cite{kutbi2021zero}. We denote paired TI data as $\{x_{TI_i}^{m_S},x_{TI_i}^{m_T}\}_{i=1}^{n_{TI}}$, where  $x_{TI_i}^{m_S}$ is the $i$-th TI data point from source modality and $x_{TI_i}^{m_T}$ is its paired counterpart from the target modality, $n_{TI}$ the total number of pairs. We compute our proposed loss $\mathcal{L}_{TI}$ using TI data as follows:

\noindent\textbf{Step 1:} We feed source modality images of the TI dataset through each of the source models to pre-compute features that are good representations of modality $m_S$. We denote the $i$-th TI source feature extracted from source $j$ as $\psi_j^i$:
\begin{equation}
    \psi_j^i = f_j(x_{TI_i}^{m_S}). 
\label{precompute_feature}
\end{equation}

\noindent\textbf{Step 2:} During the knowledge transfer phase, we feed the target modality images of the TI dataset which are encouraged to match the corresponding pre-extracted source modality features. We do so by minimizing $\mathcal{L}_{TI}$ defined below with respect to the parameters in the feature encoders for the target modality:
\begin{equation}
    \mathcal{L}_{TI} = \sum_{i=1}^{n_{TI}} \sum_{j=1}^n  
    \normx{\zeta_j (\psi_j^i - f_j(x_{TI_i}^{m_T}))}^2.
\label{L_TI}
\end{equation}

\subsection{Task-relevant distribution matching}

\label{sec:TR_stat}

In the task-irrelevant feature matching, we match the TI features of two modalities in the feature space. Even if this captures some class independent cross modal mapping between source and target modalities, it has no information about the \textit{TR-class conditional cross modal mapping}. By this term we refer to the cross modal relationship between source and target, given the relevant classes. Assuming that the marginal distribution of the source features across the batches can be modeled as Gaussian, such feature statistics can be fully characterized by its mean and variance. We propose to match the feature statistics across the source and target, to reduce the modality gap further. 

It might seem as though some amount of source data would be required to estimate the batch-wise mean and and variance of its feature map, but the running average statistics stored in the conventional BatchNorm (BN) layers are good enough to serve our purpose. The BN layers normalize the feature maps during the course of training to mitigate the covariate shifts \cite{ioffe1502accelerating,yin2020dreaming}. As a result it is able to capture the channel-wise feature statistics cumulatively over all the batches, which gives rise to a rough estimate of the expected mean and variance of the batch-wise feature map, at the end of training. Let us consider that the BN layer corresponding to the $l$-th convolution layer ($\mathcal{B}_l$) has $r_l$ nodes and there exist $b$ number of such layers per source model. Then we refer to the expected batch-wise mean and variance of the $l$-th convolution layer of the $k$-th source model as $\mathbb{E}\big[\mu_l|\mathcal{X}_{S_k}^{m_S}\big] \in \mathbb{R}^{r_l}$ and $\mathbb{E}\big[\sigma^2_l|\mathcal{X}_{S_k}^{m_S}\big] \in \mathbb{R}^{r_l}$. 

Prior to the start of the knowledge transfer phase, we pre-extract the information about the source feature statistics from all of the pre-trained source models. During the knowledge transfer phase, for each iteration we calculate the batch-wise mean and variance of the feature map of target data from all the source models, linearly combine them according to the weights $\zeta_i$ and minimize the distance of this weighted combination with the weighted combination of the pre-computed source feature statistics. We calculate this loss $\mathcal{L}_{d}$ given by
\begin{equation}
    \mathcal{L}_{d} =  \sum_{l=1}^{b}   \Bigg(
    \normx{\sum_{j=1}^n \zeta_j \mathbb{E}\big[\mu_l|\mathcal{X}_{S_j}^{m_S}\big] - \sum_{j=1}^n \zeta_j \hat{\mu}_{l_j}} + \\ 
     \normx{\sum_{j=1}^n \zeta_j \mathbb{E}\big[\sigma^2_l|\mathcal{X}_{S_j}^{m_S}\big] - \sum_{j=1}^n \zeta_j \hat{\sigma^2}_{l_j}} \Bigg),
\label{L_TR}
\end{equation}
where $\mathbb{E}\big[\mu_l|\mathcal{X}_{S_j}^{m_S}\big]$ and $\mathbb{E}\big[\sigma^2_l|\mathcal{X}_{S_j}^{m_S}\big]$ are the running mean and variance of the batchnorm layer corresponding to the $l$-th convolution layer of source $j$, which we refer as $\mathcal{B}_l^j$, and $\hat{\mu}_{l_j}= \frac{1}{n_T}\sum_{k=1}^{n_T}\mathcal{B}_l^j(x_T^k)$  and 
$\hat{\sigma^2}_{l_j}= \frac{1}{n_T}\sum_{k=1}^{n_T}(\mathcal{B}_l^j(x_T^k)-\hat{\mu}_{l_j})^2$ denote the mean and variance of the target output from the same batchnorm layer. The losses $\mathcal{L}_{TI}$ and $\mathcal{L}_{d}$ minimize the modality gap between source and target. We name the combination of these two losses as \textit{Modality Specific Loss} $\mathcal{L}_{ms} = \lambda_{TI} \mathcal{L}_{TI} + \lambda_{d} \mathcal{L}_{d}$, where $\lambda_{TI}$ and $\lambda_d$ are regularization hyper-parameters.

\subsection{Overall optimization}

The two proposed methods above help to reduce the modality gap between source and target without accessing task-relevant source data. In addition to them, we employ the unlabeled target data directly for knowledge transfer. Specifically, we perform \textit{information maximization} along with minimization of a self-supervised \textit{pseudo-label loss}, which have shown promising results in source-free UDA \cite{liang2020we,ahmed2021unsupervised} where the source and target modalities are the same.

\noindent\textbf{Information Maximization (IM)}: IM is essentially the task of performing maximization of the mutual information between distribution of the target data and its labels predicted by the source models. This mutual information is a combination of a conditional and a marginal entropy of the target label distribution. 

Motivated by \cite{ahmed2021unsupervised}, we calculate the \textit{conditional entropy} $\mathcal{L}_{ent}$ and the marginal entropy termed as \textit{diversity} $\mathcal{L}_{div}$  as follows:
\begin{equation}
    \mathcal{L}_{ent} = -\frac{1}{n_T}\Big[ \sum_{i=1}^{n_T} (\mathcal{F}_{T}^{m_T}(x_T^i)) \log(\mathcal{F}_{T}^{m_T}(x_T^i))\Big], \mathcal{L}_{div} =  -\sum_{j=1}^N \Bar{p}_j\log\Bar{p}_j,
\label{Im_ent}
\end{equation}
where $\mathcal{F}_{T}^{m_T}(x_T^i)= \sum_{k=1}^n \zeta_k \mathcal{F}_{S_k}^{m_S}(x_T^i) $, $\zeta_k$ is the weight assigned to the $k$-th source such that $\zeta_k \geq 0$ , $\sum_{k=1}^n \zeta_k = 1$ and $\Bar{p}= \frac{1}{n_T}\sum_{i=1}^{n_T}\Big[\mathcal{F}_{T}^{m_T}(x_T^i)\Big]\in \mathbb{R}^N$ is the empirical label distribution. The \textit{mutual information} is calculated as $\mathcal{L}_{IM}=\mathcal{L}_{div}-\mathcal{L}_{ent}$. Maximization of $\mathcal{L}_{IM}$ (or minimization of $-\mathcal{L}_{IM}$) ensures the target labels, as predicted by the sources, more confident and diverse in nature.

\noindent\textbf{Pseudo-label loss:} Maximizing $\mathcal{L}_{IM}$ helps to obtain labels that are more confident in prediction and globally diverse. However, that does not prevent mislabeling (\ie, assigning wrong labels to the inputs), which leads to \textit{confirmation bias} \cite{tarvainen2017mean}. To alleviate this problem, we adopt a self supervised pseudo-label based cross entropy loss, inspired by \cite{ahmed2021unsupervised,liang2020we} (see the supplement for the exact details about computing the self-supervised pseudo-labels.) After calculating pseudo-labels we compute the \textit{pseudo-label cross entropy} loss $\mathcal{L}_{pl}$ as follows:
\begin{equation}
    \mathcal{L}_{pl} = - \frac{1}{n_T}\sum_{i=1}^{n_T}\sum_{k=1}^K \mathbf{1}\{\hat{y}_T^i=k\} \log \big[\mathcal{F}_{T}^{m_T}(x_T^i)\big]_k,
\label{Im_pl}
\end{equation}
where $\hat{y}_T^i$ is the pseudo-label for the $i$-th target data point and $\mathbf{1}\{.\}$ is an indicator function that gives  value $1$ when the argument is true. Our final loss is the combination of the above two losses. We call this combination \textit{modality agnostic loss} $\mathcal{L}_{ma}$, which is expressed as $\mathcal{L}_{ma}= -\mathcal{L}_{IM}+\lambda_{pl}\mathcal{L}_{pl}$.

We calculate the overall objective function as the sum of \textit{modality agnostic} and \textit{modality specific} losses and optimize Eq.~\eqref{opt:main_opt} using Algorithm~\ref{algo1}.
\begin{mini}|l|
{\{f_j\}_{j=1}^n,\zeta}{\mathcal{L}_{ma} + \mathcal{L}_{ms}}{}{}
\text{\quad s.t.} {\sum_{k=1}^n\zeta_k = 1, \zeta_k \geq 0}
\label{opt:main_opt}
\end{mini}

\begin{algorithm}[]
\SetAlgoLined
\small
\textbf{Input:} $n$ source models trained on modality $m_S$ $\{\mathcal{F}_{S_k}^{m_S}\}_{k=1}^n=\{{g}_k \circ f_{k}\}_{k=1}^n$, unlabeled target data $\{x_{T}^i\}_{i=1}^{n_T}$ from modality $m_T$, TI cross modal pairs $\{x_{TI_i}^{m_S},x_{TI_i}^{m_T}\}_{i=1}^{n_{TI}}$, mixing weights $\{\zeta_k\}_{k=1}^n$, max number of epochs $E$, regularization parameters $\lambda_{TI}$, $\lambda_d$, number of batches $B$  \\
\textbf{Output:} Optimal adapted feature enocoders $\{f^\star_{k} \}_{k=1}^n$, mixing weights $\{\zeta_k^*\}_{k=1}^n$ \\
\textbf{Initialization}: Freeze final classification layers $\{g_k\}_{k=1}^n$, set $\zeta_k=\frac{1}{n}$ for all $k$\\
Calculate $\{\psi_j^i\}_{j=1}^n \ \forall i \in \small[1,2 \ldots,n_{TI}\small]$ using Eq.~\eqref{precompute_feature}\\
Retrieve $\mathbb{E}\big[\mu_l|\mathcal{X}_{S_j}\big]$ and $\mathbb{E}\big[\sigma^2_l|\mathcal{X}_{S_j}\big]$ for all $j$ and $l$ as in Section~\ref{sec:TR_stat}\\
\textbf{Knowledge Transfer Phase}: \\
 \For{$epoch=1$ $\textbf{to}$ $E$}{

  \For{$iteration=1$ \textbf{to} $B$}{
  
  Sample a mini batch of target data and feed it through each of the source models \\
  
  Calculate loss terms in Eq.~\eqref{L_TI},~\eqref{L_TR},~\eqref{Im_ent}, and~\eqref{Im_pl}\\
  
  Compute overall objective from Eq.~\eqref{opt:main_opt} \\
  
  Update parameters in $\{f_j\}_{j=1}^n$ and $\{\zeta_k\}_{k=1}^n$ by optimizing~\eqref{opt:main_opt} \\
  
  Make $\zeta$ non-negative by setting $\zeta_k:=1/(1+e^{-\zeta_k})$ \\
 
  Normalize $\zeta$ by setting $\zeta_k := \zeta_k/\sum_{i=1}^n \zeta_i$
  }
 
 }
 Final target model $\mathcal{F}_{T}^{m_T} = \sum_{k=1}^n \zeta_k^\star(g_k \circ f_k^\star)$
 \caption{Algorithm to Solve Eq.~\eqref{opt:main_opt}} 
 
 \label{algo1}

\end{algorithm}

\section{Experiments}
We first describe the datasets, baselines and experimental details we employ. Next, we show results of single and multi-source cross modal transfer which show the efficacy of our method. In Section \ref{CMCD} we demonstrate experimentally why source free cross modal is a much harder problem compared to cross domain knowledge transfer. We conclude this section by performing analysis on different hyperparameters.

\subsection{Datasets, baselines and experimental details} 
\label{data_details}

\noindent\textbf{Datasets:} To show the efficacy of our method we extensively test on publicly available cross-modal datasets. We show results on two RGB-D (RGB and Depth) datasets -- SUN RGB-D \cite{song2015sun} and DIML RGB+D \cite{cho2021deep}, and the RGB-NIR Scene (RGB and Near Infrared) dataset \cite{brown2011multi}. We summarize the statistics of the datasets in Table~\ref{tab:dataset_stats}. In the supplement, we provide examples from each dataset and the list of classes which we use as TI and TR data in our experiments. 

\begin{enumerate}[leftmargin=*]
\item {SUN RGB-D:} A scene understanding benchmark dataset which contains 10335 RGB-D image pairs of indoor scenes. The dataset has images acquired from four different sensors named \textit{Kinect version1 (kv1)}, \textit{Kinect version2 (kv2)}, \textit{Intel RealSense} and \textit{Asus Xtion}. We treat these four sensors as four different domains. Out of total $45$ classes, $17$ common classes are treated as TR classes and the remaining 28 classes as TI classes. To train four source models, one for each domain, we use the RGB images from the TR classes, specific to that particular domain. We treat the TR depth images from each of the domains as the target modality data.

\item {DIML RGB+D:} This dataset consists of more than $200$ indoor/outdoor scenes. We use the smaller sample dataset instead of the full dataset, which has $1500$/$500$ RGB-D pairs for training/testing distributed among $18$ scene classes. We split the training pairs into RGB and depth, and treat those two as source and target, respectively. The synchronized RGB-D frames are captured using Kinect v2 and Zed stereo camera \cite{kim2016structure,kim2017deep,kim2018deep}.

\item {RGB-NIR Scene:} This dataset consists of $477$ images from $9$ scene categories captured in RGB and Near-infrared (NIR). The images were captured using separate exposures from modified SLR cameras, using visible and NIR  \cite{brown2011multi}. We perform single source knowledge transfer from RGB to  NIR and vice versa for this dataset. For all the datasets, TR/TI split is done according to Table~\ref{tab:dataset_stats}.
\end{enumerate}

\noindent\textbf{Baseline Methods:}
The problem statement we focus on in this paper is new and has not been considered in literature before. As such, there is no direct baseline for our method. However, the closest related works are source free cross domain knowledge transfer methods that operates under both single and multi-source cases \cite{liang2020we,yang2021generalized,yang2021exploiting,yang2020casting,agarwal2022unsupervised,liang2021source, ahmed2021unsupervised}. SHOT \cite{liang2020we} and DECISION \cite{ahmed2021unsupervised} are the best-known works on single source and multi-source SFDA and we compare against only these two methods. Unlike \ours, neither of these baselines employ strategies to overcome modality differences and use only the modality-agnostic loss $\mathcal{L}_{ma}$ for training the target models. Using scene classification as the task of interest, we will show that \ours~outperforms these baselines for cross-modal knowledge transfer with no access to task-relevant source data. We provide details about the network architecture in the supplement. We note that there a few more recent works \cite{yang2021generalized,yang2021exploiting,yang2020casting,agarwal2022unsupervised,liang2021source} which have shown small improvements over SHOT, and are orthogonal to the ideas in this paper. Incorporating these improvements for SOCKET as well can be interesting and consider this future work.

\begin{table}[t]
\centering
\caption{Datasets statistics}
\resizebox{0.8\columnwidth}{!}{%
\begin{tabular}{@{}cccc@{}}
\toprule
 &SUN-RGBD~\cite{song2015sun} &RGB-NIR Scene~\cite{brown2011multi} & DIML~\cite{cho2021deep} \\
\midrule
Number of domains &4 &1 &1\\ 
Domain names &kv1,kv2,Realsense,Xtion &N/A &N/A \\
\# of TR images for source training  &1264,1234,238,2512 &204 &527\\
\# of TR unlabeled images  &1264,1234,238,2512 &204 &527\\
Number of TI paired images & 1709 &153 &1088\\
Number of TR \& TI classes & 17 \& 28 &6 \& 3 & 6 \& 12\\
% Indoor or outdoor scenes & Indoor & Both & Both\\
Modalities &RGB-D &RGB-NIR &RGB-D\\

\bottomrule
\end{tabular}%
}
\label{tab:dataset_stats}
\end{table}

\noindent\textbf{Performing knowledge transfer:} Recall that we initialize the target models with the source weights and the classifier layers are frozen. The weights in the feature encoders and source mixing weight parameters ($\zeta_k$'s) in the case of multi-source are the optimization parameters. The values of various parameters like the learning rate are given in the supplement.

$\lambda_{pl}$ is set as $0.3$ for all the experiments following \cite{ahmed2021unsupervised}. For the regularization parameters $\lambda_{TI}$ and $\lambda_{d}$ of \textit{modality specific} losses, we set them to be equal. We empirically choose those parameters in such a way so as to balance it with the \textit{modality agnostic} losses such that no loss component overpowers the other by a large margin. Empirically we found that a range of $(0.1,0.5)$ works best. All of the values in this range outperform the baselines and we report the best accuracies amongst those. For images from the modalities other than RGB, which are depth and NIR, we repeat the single-channel images into three-channel images, to be able to feed it through the feature encoders which are initialized from the source models trained on RGB images. We use a batch size of $32$ for all of our experiments. We run our method 3 times for all experiments with 3 random seeds in PyTorch \cite{paszke2019pytorch} and report the average accuracies over those.

\begin{table*}[t]
\centering
\caption{\textbf{Results on the SUN RGB-D dataset \cite{song2015sun} for the task of single-source cross-modal knowledge transfer from RGB to depth modality without access to task relevant source data.} The rows represent RGB domains on which the source models are trained. The columns represent the knowledge transfer results on the depth domains for three methods -- \textit{Unadapted} shows results with unadapted source, SHOT\cite{liang2020we} and \ours.}
\resizebox{\textwidth}{!}{
\begin{tabular}{@{}lllllllllllll@{}}
\toprule
\multirow{2}{*}{\diagbox{Source RGB}{Target depth}} & \multicolumn{3}{c}{Kinect v1}   & \multicolumn{3}{c}{Kinect v2}   & \multicolumn{3}{c}{Realsense} & \multicolumn{3}{c}{Xtion} \\ \cmidrule(l){2-13} 
                  & Unadapted & SHOT & \ours & Unadapted & SHOT & \ours & Unadapted   & SHOT  & \ours  & Unadapted & SHOT & \ours \\ \cmidrule(r){1-1}
Kinect v1               &14.8             & 16.7      & {\textbf{25.3}}      & 14.6            &20.3      &{\textbf{23.6}}      & 9.0               &11.9        &{\textbf{13.4}}       & 7.1       &15.3      &{\textbf{18.1}}           \\
Kinect v2               & 4.0            &12.8      & {\textbf{13.6}}     &17.0             & 29.4      & {\textbf{35.2}}     &10.8               & 19.3      &{\textbf{22.8}}       & {\textbf{10.6}}           & 7.0     &  8.3     \\
Realsense         &  2.0            &7.9      &{\textbf{20.3}}      & 7.1            &18.4      & {\textbf{23.5}}     & 14.7               & 27.4      & {\textbf{30.0}}      &5.1             & 9.5     & {\textbf{11.8}}     \\
Xtion             & 0.7            &9.5      &{\textbf{14.2}}     & 6.0             & 20.2     & {\textbf{24.2}}     & 9.0               & 21.8       &  {\textbf{23.5}}     & 8.1             & 13.2      & {\textbf{22.2}}      \\ \midrule
Average           & 5.4            & 11.7     & {\textbf{18.4}}      &11.2             &22.1      &{\textbf{26.6}}      & 10.9              & 20.1     & {\textbf{22.4}}       & 7.7            & 11.3      &  {\textbf{15.1}}    \\ \bottomrule
\end{tabular}}
\label{tab:single_sun}
\end{table*}
% \usepackage{booktabs}
% \usepackage{multirow}
% \usepackage{booktabs}
% \usepackage{multirow}
% \usepackage{booktabs}
% \usepackage{multirow}
% \usepackage{booktabs}
\begin{table*}[t]
\centering
\caption{\textbf{Results on the SUN RGB-D dataset \cite{song2015sun} for the task of multiple cross-modal knowledge transfer from RGB to depth modality without access to task relevant source data.} The rows show the six combinations of two trained source models on RGB data from four different domains. The columns represent the knowledge transfer results on the domain specific depth data for \textit{DECISION}\cite{ahmed2021unsupervised}, the current SOTA for multiple source adaptation without source data, and \ours}
\resizebox{\textwidth}{!}{
\begin{tabular}{@{}cllllllll@{}}
\toprule
\multicolumn{1}{l}{\multirow{2}{*}{\diagbox{Source RGB}{Target depth}}} & \multicolumn{2}{c}{Kinect v1} & \multicolumn{2}{c}{Kinect v2}         & \multicolumn{2}{c}{Realsense} & \multicolumn{2}{c}{Xtion} \\ \cmidrule(l){2-9} 
\multicolumn{1}{l}{}                  & DECISION     & \ours     & DECISION                 & \ours & DECISION        & \ours        & DECISION      & \ours      \\ \cmidrule(r){1-1}
Kinect v1 + Kinect v2                               & 17.9         & {\textbf{19.5}}      & 34.2 & {\textbf{36.6}}  & 18.8             & {\textbf{19.8}}             & 14.6              & {\textbf{18.0}}          \\
Kinect v1 + Realsense                         &  12.6            & {\textbf{18.0}}         & 23.3                         & {\textbf{26.8}}     & 24.3                & {\textbf{24.7}}            &10.9               &{\textbf{12.2}}           \\
Kinect v1 + Xtion                             & 11.7             &  {\textbf{23.9}}        & 29.6                         &{\textbf{35.7}}      & 20.3                & {\textbf{21.1}}            & 16.7              & {\textbf{20.0}}          \\
Kinect v2 + Realsense                         & 7.4             & {\textbf{11.7}}         &  22.7                        & {\textbf{33.1}}     &  28.4               &{\textbf{29.4}}             & 6.9              & {\textbf{9.1}}          \\
Kinect v2 + Xtion                             & 14.8              & {\textbf{16.2}}         & 27.0                         & {\textbf{31.0}}    & {\textbf{25.4}}                &  25.0           &  11.6             & {\textbf{18.3}}           \\
Realsense + Xtion                       & 8.3             &{\textbf{10.7}}          & 23.1                         & {\textbf{25.2}}     & 30.1                &  {\textbf{31.5}}           &9.5               &{\textbf{10.8}}           \\ \midrule
Average                               &  12.1            &  {\textbf{16.6}}        &  26.7                        & {\textbf{31.4}}     &  24.6               & {\textbf{25.3}}            & 11.7               & {\textbf{14.7}}   \\ \bottomrule      
\end{tabular}}

\label{tab:multi2_sun}

\end{table*}

\subsection{Main results}
\label{single_multi}

\noindent\textbf{Results on the SUN RGB-D dataset \cite{song2015sun}:} Our method is general enough to deal with any number of sources and we demonstrate both single and multi-source knowledge transfer. In Table~\ref{tab:single_sun}, we show single source RGB to depth results for all of the four domains. Treating the unlabeled depth data of each domain as target, we adapt these using source models trained on RGB data from each of the four domains. It is easily evident from  Table~\ref{tab:single_sun}, that for the target domains Kinect V1, Kinect V2, Realsense and Xtion, \ours~consistently outperforms the baseline by a good margin of $6.7\%$, $4.5\%$, $2.3\%$, and $3.8\%$, respectively, thus proving the efficacy of \ours~in a source-free cross modal setting. In some of the cases \ours~outperforms the baseline by a very large margin, as high as $12.4\%$ (Realsense-RGB to Kinect V1-depth). We show two-source RGB to depth adaptation results in Table~\ref{tab:multi2_sun}. For four domains we get six two-source combinations, each of which is used for adaptation to depth data from all four domains. We see that in this case also, on average \ours~outperforms the baseline for all four target domains by good margins. \ours~shows good improvement for some individual cases like (Kinect v1 + Xtion)-RGB to Kinect v1 depth -- improvement of $12.2\%$ -- and (Kinect v2 + Realsense)-RGB to Kinect v2 depth --improvement of $10.4\%$.

\begin{table}[t]
\centering
\caption{
Classification accuracy (\%) on DIML dataset with different TI data}
\resizebox{0.7\textwidth}{!}
{
\begin{tabular}{@{}ccccc@{}}
\toprule
 & Unadapted & SHOT \cite{liang2020we} &\ours &\ours \\ 
TI data &N/A &N/A &DIML RGB+D &SUN RGB-D\\
\midrule
RGB$\rightarrow$Depth & 26.9        & 41.4 &\textbf{46.1}          &\textbf{53.2}                  \\ \bottomrule
\end{tabular}} 
\label{tab:DIML}
\end{table}

\noindent\textbf{Results on the DIML RGB+D dataset \cite{cho2021deep}:} We performed a single source adaptation experiment (Table~\ref{tab:DIML}) by restructuring the dataset according to Table~\ref{tab:dataset_stats}. In Table~\ref{tab:DIML}, we use the TI data from both the DIML RGB+D as well as SUN RGB-D datasets in two separate columns, where the TI data of SUN RGB+D is the same that have been used for experiments related to the SUN RGB-D dataset. By doing so, we show that \ours~can perform well even with TI data from a completely different dataset, and find that \ours~has a gain of $4.7\%$ and $11.8\%$ over baseline for these two TI data settings, respectively.

\begin{table}[t]
\centering
\caption{Results on RGB-NIR dataset \cite{brown2011multi} for the task of single-source cross-modal knowledge transfer from RGB to NIR and vice versa without task-relevant source data}
\resizebox{0.5\textwidth}{!}
{
\begin{tabular}{@{}lccc@{}}
\toprule
\diagbox{Setting}{Method} & Unadapted & SHOT \cite{liang2020we} &\ours \\ \midrule
RGB $\rightarrow$ NIR & 84.8        & 86.7 &\textbf{90.2} \\
NIR $\rightarrow$ RGB  & 65.2        & 92.2     &\textbf{92.7}\\
\bottomrule
\end{tabular}}

\label{tab:NIR}
\end{table}

\noindent\textbf{Results on the RGB-NIR scene dataset \cite{brown2011multi}:} We now show that \ours~also outperforms baslines when the modalitiies are RGB and NIR using the RGB-NIR dataset. We follow the splits described in Table~\ref{tab:dataset_stats}. We do experiments on both RGB to NIR and vice versa. The results are given in Table~\ref{tab:NIR}. For RGB to NIR transfer, \ours~shows $3.5\%$ improvement, while for NIR to RGB transfer, it shows $0.5\%$ improvement over the competing method.

\begin{table}[t]
\centering
\caption{\textbf{Cross modal vs cross domain knowledge transfer for SUN RGB-D dataset scene classification using SHOT\cite{liang2020we}}: (1) The first columm shows the accuracies for RGB to depth transfer within the same domain. (2) The second  column is generated by transferring knowledge from one RGB domain to other three RGB domains taking the average of the accuracies}
\resizebox{0.45\columnwidth}{!}{
\begin{tabular}{lcc}
\toprule
Source & Cross-Modal & Cross-Domain \\ 
\midrule
Kinect v1       & 16.7                             & 24.5                   \\
Kinect v2       & 29.4                            & 39.6                   \\
Realsense & 27.4                          & 29.7                   \\
Xtion     & 13.2                   & 43.1                   \\
\midrule
Average   & 21.7                                    & 34.2                   \\ \bottomrule
\end{tabular}
}
\label{tab:cmcd}
\end{table}

\subsection{Cross Modal vs Cross Domain}
\label{CMCD}
In order to show the importance of the novel problem we consider, we compare the single-source knowledge transfer results on the SUN RGB-D dataset for modality change vs domain shift in Table~\ref{tab:cmcd}. We use SHOT \cite{liang2020we} which is a source-free UDA method for this experiment. All the domain-specific source models are trained on RGB images. For domain shift, the targets are all the RGB images of the remaining 3 domains and we report the average over them. Domain shift involves changes in sensor configuration, viewpoints, etc. For modality change, the target data are depth images from the same domain. The scenes are the same as in the RGB source, except they are captured using the depth sensor. The table clearly shows that the accuracy drops by a large margin of $12.5\%$ when we transfer knowledge across modalities instead of domains of the same modality. This shows that a cross-modal knowledge transfer is not the same as DA and a framework like \ours~is necessary to reduce the modality gap.

\begin{table}[t]
\centering
\caption{\textbf{Ablation of contribution of our proposed novel loss components.} The first accuracy column (a) corresponds to single source adaptation from RGB to depth on \textit{kv2} domain, whereas the second column (b) shows the multi-source adaptation result from \textit{kv1+xtion} to \textit{kv1} domain of SUN RGB-D dataset. We show the accuracy gain over using $\mathcal{L}_{ma}$ only inside the parentheses}
\resizebox{.5\columnwidth}{!}{
\begin{tabular}{@{}cccll@{}}
\toprule
$\mathcal{L}_{ma}$ &$\mathcal{L}_{d}$ &$\mathcal{L}_{TI}$ &(a) accuracy (\%) &(b) accuracy (\%)\\
\midrule
\color{ForestGreen}\cmark & & &30.0 &11.7\\
\color{ForestGreen}\cmark &\color{ForestGreen}\cmark & &31.6 ($\uparrow$1.6) &18.3 ($\uparrow$6.6)\\
\color{ForestGreen}\cmark & &\color{ForestGreen}\cmark &34.9 ($\uparrow$4.9) &22.6 ($\uparrow$10.9)\\
\color{ForestGreen}\cmark &\color{ForestGreen}\cmark &\color{ForestGreen}\cmark &\textbf{36.3} ($\uparrow$6.3) &\textbf{23.9} ($\uparrow$12.2)\\ \bottomrule
\end{tabular}
}
\label{ablation_loss}
\end{table}

\subsection{Ablation and sensitivity analysis}
\noindent\textbf{Contribution of loss components}: In Table~\ref{ablation_loss}, the first row has the result with just the \textit{modality agnostic} loss $\mathcal{L}_{ma}$, whereas second and third row shows the individual effect of our proposed \textit{modality specific} losses along with the $\mathcal{L}_{ma}$. For all cases, \ours~outperforms the baseline and using both losses in conjunction with $\mathcal{L}_{ma}$ yields best results.

\noindent\textbf{Effect of number of TI images:} We randomly chose six classes from SUN RGB-D dataset as TI data. Table~\ref{tab:sensitivity_analysis} clearly shows that increasing per class samples of TI data results in improving the scene-classification accuracy for RGB to depth transfer on the SUN RGB-D dataset. In short, for a fixed number of TI classes, the more TI images per class, the better \ours~performs.

\noindent\textbf{Effect of regularization parameters:} In Table~\ref{tab:sensitivity_analysis}, we observe the effect of test accuracy vs the regularization hyper-parameters for our novel losses proposed as a part of \ours. We keep $\lambda_{TI}$ and $\lambda_d$ equal to each other for values between $0$ to $1$. Using the value of $0$ is the same as using SHOT. From the table, we see that as the value of the parameter increases the accuracy also increases up to a certain point, and then it starts decreasing.

\begin{table}[t]
\centering
\caption{\textbf{Left: Effect of number of TI data.} We perform knowledge transfer from Kinect v1 RGB to unlabeled depth data. We use six random TI classes and vary the number of TI images per class from $0$ to $60$ in steps of 20. \textbf{Right: Effect of regularization hyper-parameters}. We perform Kinect v1 and Kinect v2 RGB to Kinect v1 depth transfer with varying $(\lambda_{TI},\lambda_d)$ and tabulate the accuracy of SOCKET}
\resizebox{.4\columnwidth}{!}{
{\begin{tabular}{@{}lllllll@{}}
\toprule
Images per class & 60   & 40   & 20  & 0     \\ \midrule
Accuracy (\%)               & 25.0 & 22.5 & 20.3 & 16.7  \\ \bottomrule
\end{tabular}}
}
\resizebox{.5\columnwidth}{!}{
\begin{tabular}{p{2cm}p{1cm}p{1cm}p{1cm}p{1cm}p{1cm}}
\toprule
$(\lambda_{TI},\lambda_d)$         & 0.00    & 0.05 & 0.10  & 0.50  & 1.00  \\ \midrule
Kinect v1 & 16.1 & 15.0  & 16.6 & 23.4 & 21.0 \\ 
Kinect v2 & 29.3 & 34.2   & 35.0 & 36.7 & 16.3 \\ \bottomrule
\end{tabular}}
\label{tab:sensitivity_analysis}
\end{table}

\section{Conclusion}

We identify the novel and challenging problem of cross-modality knowledge transfer with no access to the task-relevant data from the source sensor modality, and only unlabeled data in the target. We propose our framework, \ours, which includes devising loss functions that help bridge the gap between the two modalities in the feature space. Our results for both RGB-to-depth and RGB-to-NIR experiments show that \ours~outperforms the baselines which cannot effectively handle modality shift.

\noindent\textbf{Acknowledgements.}
SMA, SL, KCP and MJ were supported by Mitsubishi Electric Research Laboratories. SMA and ARC were partially supported by ONR grant N00014-19-1-2264 and the NSF grants CCF-2008020 and IIS-1724341.

\clearpage

\bibliographystyle{splncs}
\bibliography{egbib}

% \onecolumn
\newpage

\begin{center}
\textbf{\Large{Cross-Modal Knowledge Transfer Without Task-Relevant Source Data \\
\vspace{1cm}
(Supplementary Material)}}
\end{center}

\section{Dataset example images}

\begin{figure*}[!ht]
\centering
\includegraphics[width=.99\textwidth]{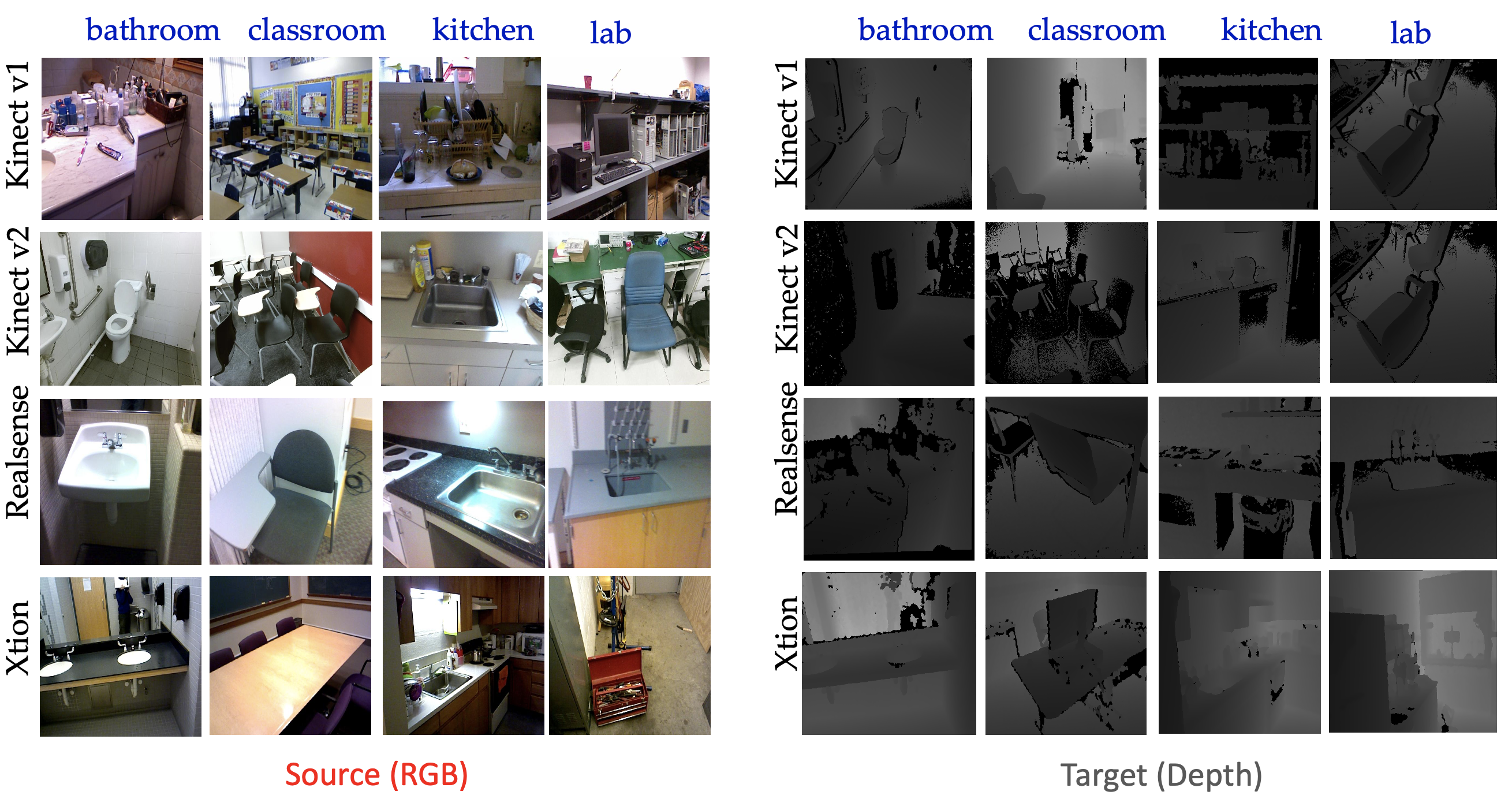} 
\caption{\textbf{SUN RGB-D TR samples.} We show some example images of the four domains of SUN RGB-D. Both modalities from $4$ out of $17$ TR classes are shown here. We discard the RGB source data after training four source models and we do not use any label information for the target depth data.}
\label{TR_imgs}
\end{figure*}

\begin{figure*}[!ht]
\centering
\includegraphics[width=.99\textwidth]{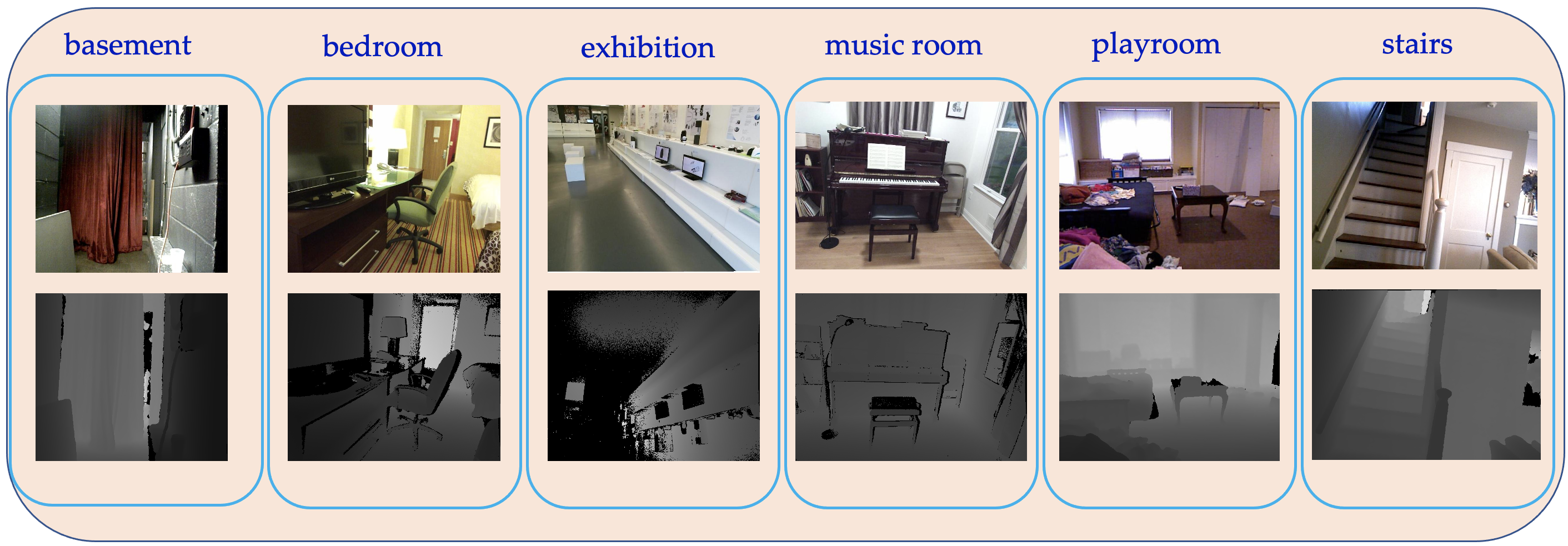} 
\caption{\textbf{SUN RGB-D TI samples.} We show some example images of the TI data from SUN RGB-D dataset. Six classes, each with paired example of RGB and depth are shown here. The TR and TI classes are completely disjoint.}
\label{TI_imgs}
\end{figure*}

\begin{figure*}[!ht]
\centering
\includegraphics[width=.99\textwidth]{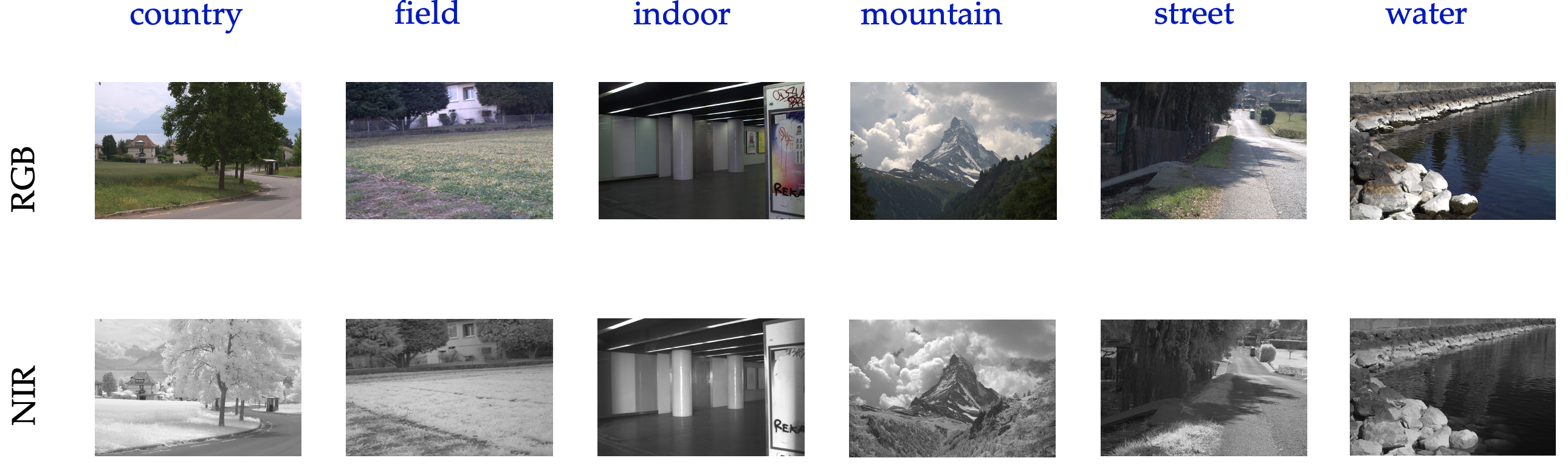} 
\caption{\textbf{RGB-NIR scene samples.} We show some example images of the  of RGB-NIR scene dataset. Both modalities of all  $6$ TR classes are shown here. We discard the source data after training the source model and we do not use any label information for the target data.}
\label{TR_nir_imgs}
\end{figure*}

\begin{figure*}[!ht]
\centering
\includegraphics[width=.6\textwidth]{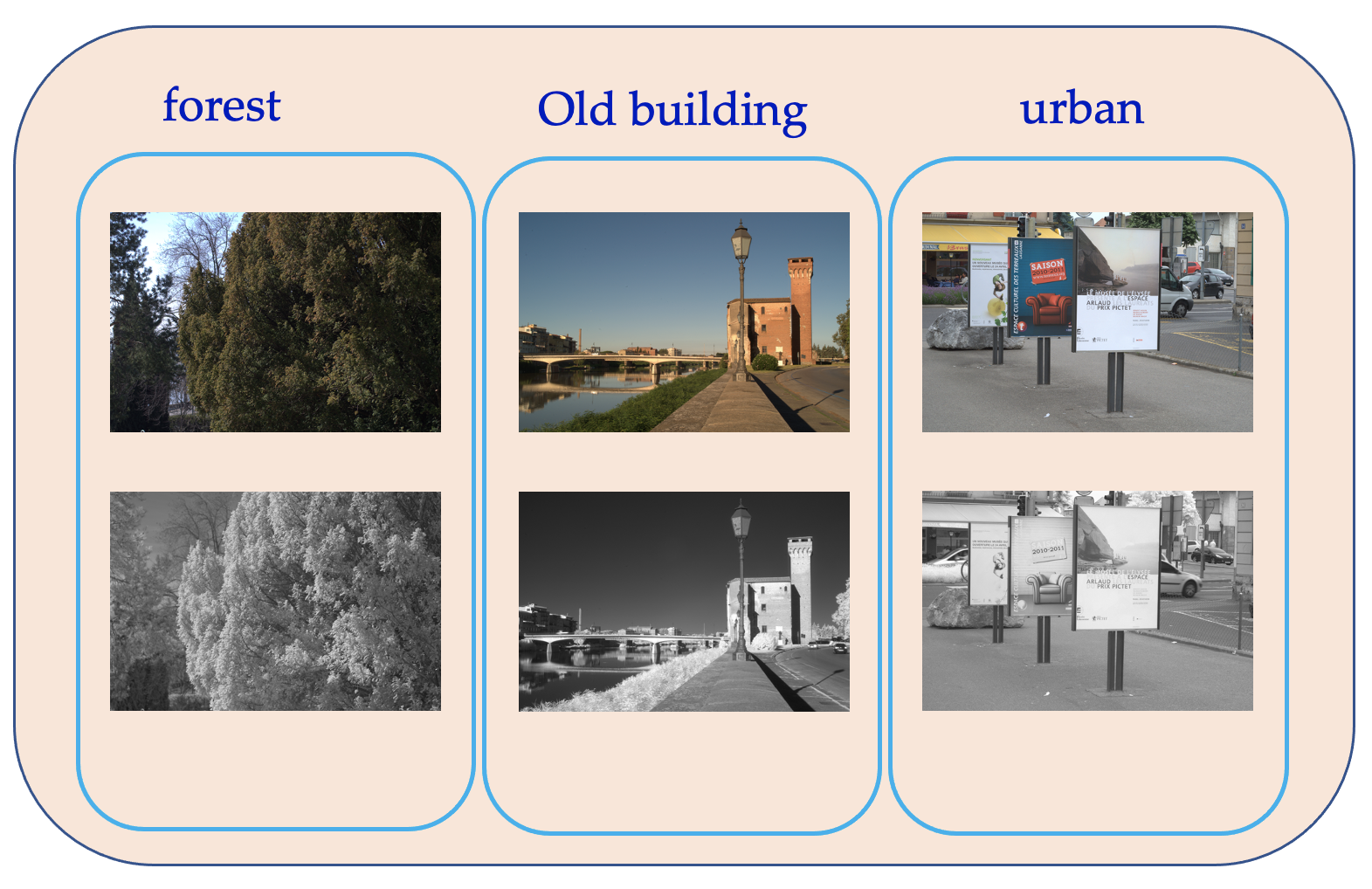} 
\caption{\textbf{RGB-NIR scene samples} We show some example images of the TI data from RGB-NIR scene dataset. Three classes, each with paired example of RGB and NIR are shown here. The TR and TI classes are completely disjoint.}
\label{TI_nir_imgs}
\end{figure*}

In Figure~\ref{TR_imgs} and Figure~\ref{TI_imgs} we show some example samples from SUN RGB-D dataset, whereas in Figure~\ref{TR_nir_imgs} and Figure~\ref{TI_nir_imgs}, samples from RGB-NIR scene dataset has been shown. For both datasets, some random samples of Task-Relevant (TR) and and Task-Irrelevant (TI) classes are shown. As DIML dataset has most of the classes overlapped with SUN RGB-D, we do not show examples for that dataset here. For the TR classes, source data are discraded after training the source models and we transfer knowledge from those models to the unlabeled data of target modality. For the TI classes, we have paired samples from both modalities. Note that for all the cases, TR and TI classes are completely disjoint.

\section{Calculation of pseudo-labels}
For these steps, we mainly follow \cite{ahmed2021unsupervised, liang2020we}. We first compute the cluster centroids of all the classes, followed by linearly combining the centroids using the current learned weight vector. We then take each of the weighted features and label it according to it's nearest neighbors from the set of $K$ weighted centroids. In the next step, we update the pseudo labels by repeating these steps. Below, we describe mathematically these steps in detail: \\

\begin{enumerate}
    \item We first compute the cluster centroids of all the classes $k \in \{1,2,\ldots N\}$ induced by source $j \in \{1,2,\ldots,n\}$ for the $0$-th iteration, by the following equation:
    \begin{equation}
     c_{k_j}^{(0)} = \frac{\sum_{x_T \in \mathcal{X}_T^{m_T}} \big[\Tilde{\mathcal{F}}_{S_j}^{m_S}(x_T)\big]_{k} \Tilde{f}_j(x_T)}{\sum_{x_T \in \mathcal{X}_T^{m_T}} \big[\Tilde{\mathcal{F}}_{S_j}^{m_S}(x_T)\big]_{k}}
    \end{equation}
    where $\big[.\big]_k$ indicates the $k$-th element of the vector in argument , $\Tilde{f}_j$ denotes the $j$-th source model's feature extractor and $\Tilde{\mathcal{F}}_{S_j}^{m_S}= g_j \circ \Tilde{f}_j$ represents the complete $j$-th source model from the last iteration. 
    \item In the next step, we linearly combine these centroids as well as the target features extracted from all the source models from last iteration, with the current learned weight vector $\zeta$ as follows: 
\begin{equation}
    c_k^{(0)} = \sum_{j=1}^n \zeta_j c_{k_j}^{(0)}
\end{equation}
\begin{equation}
  \Bar{x}_T = \sum_{j=1}^n \zeta_j \Tilde{f}_j(x_T)
\end{equation}
\item We take each of the weighted features and label it according to it's nearest neighbour from the set of $K$ weighted centroids, \ie, for a particular target feature, if the nearest neighbour is $k$-th centroid, we assign class label $k$ for that particular feature. The assigned pseudo-label $\hat{y}_T^{i(0)}$ for the $i$-th target feature $\Bar{x}_T^i$ at iteration $0$ is calculated as:
\begin{equation}
    \hat{y}_{T}^{i(0)} = \text{arg} \ \underset{k}{\text{min}} \ \|\Bar{x}_T^i-c_k^{(0)}\|_2^2
\end{equation}
\item We update the pseudo-labels in the next iteration by repeating the steps as follows:
\begin{equation}
     c_{k_j}^{(1)} = \frac{\sum_{x_T \in \mathcal{X}_T^{m_T}} \mathbf{1}\{\hat{y}_T^{(0)}=k\} \Tilde{f}_j(x_T)}{\sum_{x_T \in \mathcal{X}_T^{m_T}} \mathbf{1}\{\hat{y}_T^{(0)}=k\} }
    \end{equation}
where, $\mathbf{1}\{.\}$ is an indicator function operator which takes value $1$, when its argument is true.
\begin{equation}
    c_k^{(1)} = \sum_{j=1}^n \zeta_j c_{k_j}^{(1)}
\end{equation}
\begin{equation}
    \hat{y}_{T}^{i(1)} = \text{arg} \ \underset{k}{\text{min}} \ \|\Bar{x}_T^i-c_k^{(1)}\|_2^2
\label{pseudo1}
\end{equation}
Following the protocol of \cite{ahmed2021unsupervised}, we take $\hat{y}_{T}^{(1)}$ as the final pseudo-label $\hat{y}_{T}^{i}$, without further reiteration.

Finally the \textit{pseudo-label cross entropy} loss $\mathcal{L}_{pl}$ is calculated as follows:
\begin{equation}
    \mathcal{L}_{pl} = - \frac{1}{n_T}\sum_{i=1}^{n_T}\sum_{k=1}^K \mathbf{1}\{\hat{y}_T^i=k\} \log \big[\mathcal{F}_{T}^{m_T}(x_T^i)\big]_k.
\label{Im_pl_supp}
\end{equation}
\end{enumerate}

\section{More details about datasets}

\noindent\textbf{SUN RGB-D}\cite{song2015sun}: The $17$ common scene classes shared among the four domains are
\textit{bathroom, classroom, computer room, conference room, corridor, discussion area, home office, idk, kitchen, lab, living room, office, office kitchen, printer room, reception room, rest space, study space}.

The $28$ scene classes used as TI data are
\textit{basement, bedroom, book store, cafeteria, coffee room, dancing room, dinette, dining area, dining room, exhibition, furniture store, gym, home, study, hotel room, indoor balcony, study space, laundromat, lecture theatre, library, lobby, mail room, music room, office dining, play room, reception, recreation room, stairs, storage room}.

\noindent\textbf{DIML RGB+D}\cite{cho2021deep}: The $6$ scene classes  used as TR data are
\textit{bathroom, classroom, computer room, kitchen, corridor, living room}.

The $12$ scene classes used as TI data are
\textit{bedroom, billiard hall, book store, cafe, church, hospital, laboratory, library, metting room, restaurant, store, warehouse}. 

\noindent\textbf{RGB-NIR Scene}\cite{brown2011multi}: The $6$ scene classes used as TR data are
\textit{country, field, indoor, mountain, street, water}.

The $3$ scene classes used as TI data are
\textit{forest, old building, urban}.

\section{Effect of regularization parameters} 

For the single source adaptation results, we empirically observe that, $(\lambda_{TI},\lambda_d)= (0.5,0.5), (0.5,0.5),(0.1,0.1), (0.5,0.5)$ yields best result for Kinect v1, Kinect v2, Realsense and Xtion as targets respectively. For the DIML RGB+D datset, the parameters are set to be $(0.5,0.5)$, whereas for the RGB-NIR scene dataset, it is set as $(0.01,0.05)$. Note that, for all of the cases this hyper-parameters are chosen to balance the two loss terms. Our method always performs better than the baseline in the range of values of the hyperparameters we tested and are close to the best accuracies reported in the paper.

\section{Network architectures}
In our experiments, we take the Resnet50 \cite{he2016deep} model pretrained on ImageNet as the backbone architecture for training the source models, the same way as \cite{xu2019larger,peng2019moment,liang2020we}. Following the architectures used in \cite{ganin2015unsupervised,ahmed2021unsupervised}, we replace the last fully connected (FC) layer with a bottleneck layer containing 256 units, within which we add a Batch Normalization \cite{ioffe2015batch} (BN) layer at the end of the FC layer. A task specific FC layer with weight normalization \cite{salimans2016weight} is added at the end of the bottleneck layer. 
\section{Training source models}
For training the source models, we resize all the source images to $224\times224$. Moreover, to increase model robustness, we use smooth labels instead of one-hot encodings \cite{szegedy2016rethinking,muller2019does} during this procedure. We set the maximum number of training epochs to $20$ for all of the sources, irrespective of the datasets. We utilize stochastic gradient descent with a momentum  $0.9$ and weight decay $10^{-3}$. The learning rates are set to $10^{-3}$ for the feature encoders ($f_k$'s) and $10^{-2}$ for the added bottleneck layer. 
During adaptation and knowledge transfer to the target modality, a learning scheduler setting similar to \cite{ganin2015unsupervised,liang2020we} $\theta = \theta_0(1+10p)^{-\frac{3}{4}}$ is used, where $\theta$ and $\theta_0$ represent the current and initial learning rates and $p$ is a real number between $0$ to $1$ which captures the training progress. $\theta_0$ is set to be $10^{-3}$ for the feature encoders ($f_k$'s) and $10^{-2}$ for the added bottleneck layers along with the source mixing weight parameters ($\zeta_k$'s). The maximum number of epochs during target adaptation is set to be $15$.

\section{Knowledge transfer details}

During adaptation and knowledge transfer to the target modality, a learning scheduler setting similar to \cite{ganin2015unsupervised,liang2020we} $\theta = \theta_0(1+10p)^{-\frac{3}{4}}$ is used, where $\theta$ and $\theta_0$ represents the current and initial learning rates and $p$ is real number between $0$ to $1$ which captures the training progress. $\theta_0$ is set to be $10^{-3}$ for the feature encoders ($f_k$'s) and $10^{-2}$ for the added bottleneck layers along with the source mixing weight parameters ($\zeta_k$'s). The learning rate decreases exponentially during the course of training. The maximum number of epochs during target adaptation is set to be $15$. 
 
\section{Modification of our algorithm in presence of TI \textit{unpaired} data}

In this section, we explore the scenario of inaccessibility of pairwise cross-modal data for TI classes. In practical scenario, one might not be able to acquire cross modal paired data. In this case we show that adversarial matching between two cross modal distributions works reasonably well. Inspired from \cite{tzeng2017adversarial}, we propose the following loss function in order to align the two cross modal data distributions which are unpaired. For this purpose, we incorporate a discriminator $\mathcal{D}$ in our framework. \\
Our adversarial loss has two components: (1) \textit{True Discriminator loss } $\mathcal{L}_{TD}$ and (2) \textit{Adversarial Discriminator loss }
$\mathcal{L}_{AD}$. The first loss tries to distinguish between source and target, while the second loss is a proxy for the generator part of the well known usual adversarial loss component, which tries to fool the discriminator in such a way, so that it fails to distinguish between source and target domain. The generator is irrelevant in our framework since we are not generating any new samples, rather as a proxy of the generator we use the same discriminator as an adversary in the second loss. In short, the first loss tries to correctly classify the source and target samples, while the second loss tries to do the opposite. Now, we describe the losses mathematically below:\\

\begin{equation}
    \mathcal{L}_{TD} = -\frac{1}{n_{TI}}\sum_{i=1}^{n_{TI}}\Bigg[\log \mathcal{D}\Big(\sum_{j=1}^n \zeta_j \psi_j^i\Big)+ \log \Big(1- \mathcal{D}\Big(\sum_{j=1}^n \zeta_j f_j(x_{TI_i}^{m_T})\Big)\Big) \Bigg]
    \label{LTD}
\end{equation}

\begin{equation}
    \mathcal{L}_{AD} = -\frac{1}{n_{TI}}\sum_{i=1}^{n_{TI}}\Bigg[\log \mathcal{D}\Big(\sum_{j=1}^n \zeta_j f_j(x_{TI_i}^{m_T})\Big)\Bigg]
    \label{LAD}
\end{equation}

Note that, $\mathcal{L}_{TD}$ is essentially a cross entropy loss computed with source TI labels as 1 and target TI labels as 0, while $\mathcal{L}_{AD}$ is also a cross entropy loss but computed with target TI labels as 1. So, clearly $\mathcal{L}_{AD}$ will try to oppose the loss $\mathcal{L}_{TD}$, so that the source and target features are indistinguishable. So our overall adversarial loss $\mathcal{L}_{adv}$ is calculated as follows:

\begin{equation}
    \mathcal{L}_{adv} = \mathcal{L}_{TD} + \lambda_{AD} \mathcal{L}_{AD}
\end{equation}
where $\lambda_{AD}$ is a regularization parameter to balance the two adversarial loss components. In the absence of TI paired data, the overall new objective function $\mathcal{L}_{tot}$ will be
\begin{equation}
    \mathcal{L}_{tot} = \mathcal{L}_{ma} + \lambda_{adv} \mathcal{L}_{adv}+\lambda_{d} \mathcal{L}_{d}
    \label{adv}
\end{equation}
To show the effectiveness of this loss, we conduct a small experiment in table \ref{adversarial_loss}. We transfer knowledge from the kv2 RGB model to unlabeled kv2 depth data. Due to time constraint we just run this algorithm with one random seed. $\lambda_{AD}$ is set to be 10 to give slightly more importance to $\mathcal{L}_{AD}$ compare to $\mathcal{L}_{TD}$, since our ultimate goal is to learn a feature embedding that can not distinguish between source and target. Clearly we see that our new adversarial loss has an increment of almost 2.9\% when used with  $\mathcal{L}_{ma}$. Though this gain is not as high compare to the case of having paired TI data (see table \textcolor{red}{8} in main paper), it is still significant and has great potential. This result is intuitively expected and show that even if with unpaired TI data, we can reduce the modality gap in the absence of TR source data. We hypothesize that for the unpaired TI data case, it is possible to reach a certain extent of the level of performance when using paired TI data, by using relatively more amount of unpaired data. We will explore it in detail for the future work.

\begin{table}[t]
\centering
\caption{\textbf{Effect of our proposed adversarial loss component.} The accuracy column corresponds to single source adaptation from RGB to depth on \textit{kv2} domain of SUN RGB-D dataset. We show the accuracy gain over using $\mathcal{L}_{ma}$ only inside the parentheses}
\resizebox{.5\columnwidth}{!}{
\begin{tabular}{@{}cccl@{}}
\toprule
$\mathcal{L}_{ma}$ &$\mathcal{L}_{d}$ &$\mathcal{L}_{adv}$ &(a) accuracy (\%) \\
\midrule
\color{ForestGreen}\cmark & & &31.0 \\
% \color{ForestGreen}\cmark &\color{ForestGreen}\cmark & &31.6 ($\uparrow$1.6) \\
\color{ForestGreen}\cmark & &\color{ForestGreen}\cmark &33.9 ($\uparrow$2.9) \\
\color{ForestGreen}\cmark &\color{ForestGreen}\cmark &\color{ForestGreen}\cmark &\textbf{34.2} ($\uparrow$3.2) \\ \bottomrule
\end{tabular}
}
\label{adversarial_loss}
\end{table}

\section{Future work, limitations and potential negative impact} 

Further studies are required to better understand the effect of amount of TI data and the diversity present in the data on the knowledge transfer results, which will require access to larger and more diverse datasets. Another interesting avenue for future  direction is applying these ideas to other modalities like point clouds, medical imaging, etc.

The work in this paper is a general method for improving knowledge transfer from a source modality to a target modality with unlabeled data. The impact of this line of research is to make it easier to train networks for modalities and tasks where large amounts of data and labeled data are not available. This may lead to a wider deployment of deep learning for such modalities. For example in applications like person re-identification, one might have access to the source models trained on private IR labeled data, which they can use to adapt RGB unlabeled data using our method, in order to match people across cameras. Thus, these algorithms can of course be good or bad for society depending on the particular application in which these ideas are employed, the bias in the datasets being used etc. This is also in true in general for other source-free DA methods \cite{liang2020we, ahmed2021unsupervised}. Therefore, steps need to be taken to ensure positive and fair outcomes of this technology.

\end{document}